\documentclass[a4paper,reprint,amsmath,amssymb,longbibliography]{revtex4-1}
\pdfoutput=1

\usepackage{amsfonts}
\usepackage{amsthm}
\usepackage{graphicx}
\usepackage{xspace}
\usepackage[spacing]{microtype}

\providecommand{\comment}[1] {}
\newcommand\iftwocol[2]{#1}

\makeatletter
\def\section{%
  \@startsection
    {section}%
    {1}%
    {\z@}%
    {-0.8cm \@plus-1ex \@minus -.2ex}%
    {0.5cm}%
    {\normalfont\small\bfseries\centering}%
}%
\def\subsection{%
  \@startsection
    {subsection}%
    {2}%
    {\z@}%
    {-.6cm \@plus-1ex \@minus -.3ex}%
    {.3cm}%
    {\normalfont\small\bfseries\centering}%
}%
\makeatother

\def\AR{\mathrm{AR}}
\def\MA{\mathrm{MA}}
\def\order{N}
\def\ac{\gamma}
\def\arc{\psi}
\def\mac{b}
\def\vX{\vec{X}}
\def\tran{^\top}
\def\innvar{\sigma}
\def\specint#1{\frac{1}{2\pi}\int_{-\pi}^\pi #1{S(\omega)} \dd \omega}
\def\fid#1{#1}

\def\cexp#1#2{#2 e^{i\omega#1}}

\def\TheTitle{Predictive Information Rate in Discrete-time Gaussian Processes}
\def\TheAbstract{%
	We derive expressions for the predicitive information rate
	(PIR) for the class of autoregressive Gaussian processes 
	$\AR(\order)$, both in terms of the prediction coefficients and in terms
	of the power spectral density. The latter result
	suggests a duality between 
	the PIR and the multi-information rate for processes with mutually inverse power
	spectra (i.e. with poles and zeros of the transfer function exchanged).
	We investigate the behaviour of the PIR in relation to 
	the multi-information rate for some simple examples, which
	suggest, somewhat counter-intuitively, that the PIR is maximised for very
	`smooth' AR processes whose power spectra have multiple poles at zero frequency.%
	We also obtain results for moving average Gaussian processes which
	are consistent with the duality conjectured earlier. One consequence of this
	is that the PIR is unbounded for $\MA(\order)$ processes.
}

\def\TheAcknowledgments{%
This research was supported by EPSRC grant EP/H01294X/1: 
`Information and neural dynamics in the perception of musical structure'.
}
\usepackage[detbar]{tools}
\usepackage{tikz}
\usetikzlibrary{calc}
\usetikzlibrary{matrix}
\usetikzlibrary{patterns}
\usetikzlibrary{arrows}
\usepackage{pgfplots}

\newcommand{\colfig}[2][1]{\includegraphics[width=#1\linewidth]{#2}}%

\newcommand\x{\vec{x}}

\def\ev(#1=#2){#1\!\!=\!#2}

\def\bet(#1,#2){#1..#2}

\newcommand{\past}[1]{\loarrow{#1}}
\newcommand{\fut}[1]{\roarrow{#1}}

\def\len{\ell}
\def\oflen{_\len}
\def\uplen{^\len}

\begin{document}
	\title{\TheTitle}
	\date{\today}
	\author{Samer A. Abdallah}
	\author{Mark D. Plumbley}
	\affiliation{Queen Mary University of London}

	\begin{abstract}
		\TheAbstract
	\end{abstract}
	\maketitle

\section{Introduction}

The \emph{predictive information rate} (PIR) of a bi-infinite 
	sequence of random variables
	$(\ldots,X_{-1},X_0,X_1,\ldots)$, was defined by \citet{AbdallahPlumbley2009}
	in the context of \emph{information dynamics}, which is concerned with the
	application of information theoretic methods \cite{Shannon48,CoverThomas} to the process of sequentially
	observing a random sequence while maintaining a probabilistic description
	of the expected future evolution of the sequence. An observer
	in this situation can maintain an estimate of its uncertainty about future
	observations (by computing various entropies) and can also estimate the information
	in each observation about the as-yet unobserved future given the all the observations
	so far; this is the \emph{instantaneous predictive information} or IPI. For stationary
	processes, the ensemble average of the IPI is the PIR. It
	is a measure of temporal structure
	that characterises the process as a whole, rather than on a 
	moment-by-moment basis or for particular realisations of the process, in the same
	way that the entropy rate characterises its overall randomness. \citet{AbdallahPlumbley2012}
	examined several process information measures and their interrelationships.
	Following the conventions established there, we let
	$\past{X}_t=(\ldots,X_{t-2},X_{t-1})$ denote the variables
	before time $t$, 
	and $\fut{X}_t = (X_{t+1},X_{t+2},\ldots)$ denote
	those after $t$.
	For a process with a shift-invariant probability measure $\mu$,
	the predictive information rate $b_\mu$ is defined as a conditional mutual information
	\begin{equation}
		\label{eq:PIR}
		b_\mu = I(X_t;\fut{X}_t|\past{X}_t) = H(\fut{X}_t|\past{X}_t) - H(\fut{X}_t|X_t,\past{X}_t).
	\end{equation}
	Equation \eqrf{PIR} says that the PIR is the average reduction
	in uncertainty about the future on learning $X_t$, given the past. 
	In similar terms, three other process information measures can be defined: the entropy
	rate $h_\mu$, the multi-information rate $\rho_\mu$ and the erasure or residual entropy rate
	$r_\mu$, as follows: 
	\begin{align}
		\label{eq:mir}
		h_\mu &= H(X_t|\past{X}_t),
\\		\rho_\mu &= I(X_t;\past{X}_t) = H(X_t) - H(X_t|\past{X}_t),
\\		r_\mu &= H(X_t|\past{X}_t,\fut{X}_t).
	\end{align}
	These measures are illustrated in an \emph{information diagram}, or I-diagram \citep{Yeung1991}, 
	in \figrf{predinfo-bg}, which shows how they partition the marginal entropy $H(X_t)$,
	the uncertainty about a single observation in isolation; this partitioning is discussed
	in depth by \citet{JamesEllisonCrutchfield2011}.

 	\begin{fig}{predinfo-bg}
		\newcommand\subfig[2]{\shortstack{#2\\[0.5em]#1}}
		\newcommand\rad{1.8em}%
		\newcommand\ovoid[1]{%
			++(-#1,\rad) 
			-- ++(2 * #1,0em) arc (90:-90:\rad)
 			-- ++(-2 * #1,0em) arc (270:90:\rad) 
		}%
		\newcommand\axis{2.75em}%
		\iftwocol{%
			\newcommand\olap{0.85em}%
			\newcommand\offs{3.6em}
			\newcommand\colsep{\hspace{5em}}
		}{%
			\newcommand\olap{0.9em}%
			\newcommand\offs{3.65em}
			\newcommand\colsep{\hspace{3em}}
		}%
		\newcommand\longblob{\ovoid{\axis}}
		\newcommand\shortblob{\ovoid{1.75em}}
		\begin{tabular}{c@{\colsep}c}
				\begin{tikzpicture}
					\newcommand\rc{2.1em}
					\newcommand\throw{2.5em}
					\coordinate (p1) at (210:1.5em);
					\coordinate (p2) at (90:0.7em);
					\coordinate (p3) at (-30:1.5em);
					\newcommand\bound{(-7em,-2.6em) rectangle (7em,3.0em)}
					\newcommand\present{(p2) circle (\rc)}
					\newcommand\thepast{(p1) ++(-\throw,0) \ovoid{\throw}}
					\newcommand\future{(p3) ++(\throw,0) \ovoid{\throw}}
					\newcommand\fillclipped[2]{%
						\begin{scope}[even odd rule]
							\foreach \thing in {#2} {\clip \thing;}
							\fill[black!#1] \bound;
						\end{scope}%
					}%
					\fillclipped{30}{\present,\future,\bound \thepast}
					\fillclipped{15}{\present,\bound \future,\bound \thepast}
					\draw \future;
					\fillclipped{45}{\present,\thepast}
					\draw \thepast;
					\draw \present;
					\node at (barycentric cs:p2=1,p1=-0.17,p3=-0.17) {$r_\mu$};
					\node at (barycentric cs:p1=-0.4,p2=1.0,p3=1) {$b_\mu$};
					\node at (barycentric cs:p3=0,p2=1,p1=1.2) [shape=rectangle,fill=black!45,inner sep=1pt]{$\rho_\mu$};
					\path (p2) +(140:3em) node {$X_0$};
					\path (p3) +(3em,0em) node  {\shortstack{infinite\\future}};
					\path (p1) +(-3em,0em) node  {\shortstack{infinite\\past}};
					\path (p1) +(-4em,\rad) node [anchor=south] {$\ldots,X_{-1}$};
					\path (p3) +(4em,\rad) node [anchor=south] {$X_1,\ldots$};
				\end{tikzpicture}%
		\end{tabular}
		\caption{
		I-diagram representation of several information measures for
		stationary random processes. Each circle or oval represents a random
		variable or sequence of random variables relative to time $t=0$. 
		The circle represents the `present'. Its total area is
		$H(X_0)=\rho_\mu+r_\mu+b_\mu$, where $\rho_\mu$ is the multi-information
		rate, $r_\mu$ is the residual entropy rate, and $b_\mu$ is the predictive
		information rate. The entropy rate is $h_\mu = r_\mu+b_\mu$.
		} \end{fig}

\section{Gauss-Markov processes}
\label{s:ar}

	A Gauss-Markov, or autoregressive Gaussian process of order $\order$ is a real-valued random process
	$(X_t)_{t\in\integers}$ on the domain of integers such that 
	\begin{equation}
		X_t = U_t + \sum_{k=1}^\order \arc_k X_{t-k},
		\label{eq:iir}
	\end{equation}
	where the \emph{innovations} $U_t$ form a sequence of independent and identically distributed
	Gaussian random variables with zero mean and variance $\innvar^2$, 
	and the $\arc_k$ are the autogressive
	or prediction coefficients.
	Thus, a realisation of the random process $X$ is the result of applying an order-$\order$
	infinite impulse response (IIR) filter to a realisation of the innovation sequence formed
	by the $U_t$. 
	The class of such processes is known as $\AR(\order)$.
	If the autregressive coefficients $\arc_k$ are such that the 
	filter is stable, the process will be stationary and thus may have well
	defined entropy and predictive information rates. We will assume that this
	is the case.
	
	\begin{fig}{ar-gm}
		\begin{tikzpicture}[->]
			\def\cn(#1,#2) {\node[circle,draw,inner sep=0.2em] (#1#2) {$#1_#2$};}
			\def\dn(#1) {\node[circle,inner sep=0.2em] (#1) {$\cdots$};}
			\def\rl(#1,#2) {\draw (#1) -- (#2);}
			\def\cl(#1,#2) {\draw (#1) to [bend right,looseness=1.2] (#2);}
			\matrix[row sep=1.3em, column sep=1.9em]{
				\dn(UH) & \cn(U,1) &  \cn(U,2) & \cn(U,3) & \cn(U,4) &  \cn(U,5) & \dn(UT) \\
				\dn(XH) & \cn(X,1) & \cn(X,2) & \cn(X,3) & \cn(X,4) & \cn(X,5) & \dn(XT) \\
			};
			\rl(XH,X1) \cl(XH,X2)
			\rl(U1,X1) \rl(X1,X2) \cl(XH,X2)
			\rl(U2,X2) \rl(X2,X3) \cl(X1,X3)
			\rl(U3,X3) \rl(X3,X4) \cl(X2,X4)
			\rl(U4,X4) \rl(X4,X5) \cl(X3,X5)
			\rl(U5,X5) \rl(X5,XT) \cl(X4,XT)
		\end{tikzpicture}
		\caption{Graphical model for a $2^\text{nd}$ order Gauss-Markov, or $\AR(2)$, process. 
		The $X_t$ are the observed, real-valued random variables, while the $U_t$ are the unobserved 
		innovations. Each $X_t$ is a deterministic (linear) function of its parents.}
	\end{fig}

	\subsection{Entropy rate}
	From the defining equation \eqrf{iir} we can immediately see that
	$H(X_t|\past{X}_t) = H(X_t|X_{t-\order},\ldots,X_{t-1}) = H(U_t)$, which depends only
	on $\innvar^2$, so 
	\begin{equation}
		h_\mu = \half \log 2\pi e \innvar^2.
		\label{eq:entro-ar}
	\end{equation}
	It is known that the entropy rate of a stationary Gaussian process can also be
	expressed in terms of its power spectral density (PSD) function $S:\reals\to\reals$,
	which is defined as the
	discrete-time Fourier transform of the autocorrelation sequence $\ac_k = \expect{X_t X_{t-k}}$:
	\begin{equation}
		S(\omega) = \sum_{k=-\infty}^\infty \ac_k e^{-i\omega k},
		\quad
		\ac_k = \specint{\cexp{k}}.
		\label{eq:psd}
	\end{equation}
	Using methods of toeplitz matrix analysis \citep{Gray2006}, the entropy rate
	can be shown, with suitable restrictions on the autocorrelation sequence, to be
	\begin{equation}
		h_\mu = \frac{1}{2} \left( \log 2\pi e  + \specint{\log} \right),
		\label{eq:kge}
	\end{equation}
	which is also known as the Kolmogorov-Sinai entropy for this process.
	Incidentally, this means that the variance of the innovations $\innvar^2$ can be expressed
	as
	\begin{equation}
		\innvar^2 = \exp \left( \specint{\log} \right).
		\label{eq:inn-var-psd}
	\end{equation}

	\subsection{Multi-information rate}
	The multi-information rate of a stationary
	Gaussian process was found by \citet{Dubnov2006} to be expressible as
	\begin{equation}
		\rho_\mu = \frac{1}{2} \left( \log \specint{\fid} - \specint{\log}\right).
		\label{eq:mir-sfm}
	\end{equation}
	We can see how this was obtained by noting that 
	the marginal variance $\expect{X_t^2} = \ac_0$
	can be computed from the spectral density function simply by setting $k=0$ in the
	inverse Fourier transform \eqrf{psd}, yeilding
	\begin{equation}
		H(X_t) = \frac{1}{2}\left( \log 2\pi e + \log \specint{}\right).
	\end{equation}
	Since $\rho_\mu = H(X_t) - h_\mu$ and $h_\mu$ is given by the Kolmogorov-Sinai
	entropy of the Gaussian process \eqrf{kge}, 
	Dubnov's expression \eqrf{mir-sfm} follows directly.

	\subsection{Predictive information rate}
	\label{s:pir}

	To derive an expression for the predictive information rate $b_\mu$,
	we first note that the model is \nth{\order}-order Markov,
	since 
	the trailing segment
	of the process $(X_{\order+1},X_{\order+2},\ldots)$ is conditionally independent of the leading
	segment $(\ldots,X_{-1},X_0)$ given intervening segment $(X_1,\ldots,X_\order)$.  
	Writing $X_{p:q}$ for the finite segment $(X_p,X_{p+1},\ldots,X_q)$,
	it can be shown that
	\begin{equation}
		I(X_0;\fut{X}_0|\past{X}_0) = I(X_0;X_{1:\order}|X_{-\order:-1}).
		\label{eq:PIR-markov}
	\end{equation}
	Thus, to find the PIR, we need only consider the 
	the $2\order+1$ consecutive variables around $X_0$, namely $X_{-\order:\order}$.
	Expanding the conditional mutual information in terms of entropies, we obtain
	\begin{equation}
		\iftwocol{
			\begin{split}
				b_\mu =
					H(X_{1:\order}|X_{-\order:-1}) - H(X_{1:\order}|X_{-\order:0}).
			\end{split}
		}{
			I(X_0;X_{1:\order}|X_{-\order:-1}) = H(X_{1:\order}|X_{-\order:-1}) - H(X_{1:\order}|X_{-\order:0}).
		}
	\end{equation}
	Since 
	the segment $X_{-\order:0}$ contains
	more than $\order$ elements, the second term is just $\order$ times the entropy rate, so
	\begin{equation}
		b_\mu = H(X_{1:\order}|X_{-\order:-1}) - \order h_\mu.
		\label{eq:pir-markov-entro}
	\end{equation}
	To evaluate $H(X_{1:\order}|X_{-\order:-1})$, we note that, for continuous random variables 
	$Y$ and $Z$,
	\begin{equation}
		H(Y|Z) = \int H(Y|\ev(Z=z)) p(z) \dd z,
	\end{equation}
	where $p(z)$ is $Z$'s probability density at $z$;
	that is, we find the entropy of $Y$ given \emph{particular}
	values of $Z$, and then average over the possible values of $Z$.
	If we find that $H(Y|\ev(Z=z))$ is the same value independent of $z$, then
	$H(Y|Z)$ is trivially that value. This is indeed what we will find when we apply
	this approach here, and so we will examine the case where the variables
	$(X_{-\order},\dots,X_{-1})$ have been observed with the values $(x_{-\order},\ldots,x_{-1})$
	respectively. 
	Under these conditions, we can, in effect, forget that $X_{-\order:-1}$ are random variables
	and investigate the joint distribution of $X_{0:\order}$ implicitly conditioned
	on the observation $\ev(X_{-\order:-1}=x_{-\order:-1})$.
	Referring back to \eqrf{iir}, we may rewrite the recursive relation between random variables 
	$X_j$ for $1\leq j \leq \order$ given observations $x_{-\order:-1}$ as
	\begin{equation}
		X_j 
			= U_j + \left( \sum_{i=1}^{j-1} \arc_i X_{j-i} \right)
						+ \arc_j X_0 
						+ \left( \sum_{i=j+1}^\order \arc_i x_{j-i} \right),
		\label{eq:cond-iir}
	\end{equation}
	with a special case for $X_0$:
	\begin{equation}
		X_0 = U_0 + \left( \sum_{i=1}^\order \arc_i x_{-i} \right).
		\label{eq:cond-iir0}
	\end{equation}
	With an eye to the final sums in both the above equations, let us also define
	$m_j$ for $0 \leq j \leq \order$ as
	\begin{equation}
		m_j = \sum_{i=j+1}^\order \arc_i x_{j-i}.
		\label{eq:offsets}
	\end{equation}
	Consider now the following transformation of variables:
	\begin{equation}
		Y_j = X_j
				- \left( \sum_{i=1}^{j-1} \arc_i X_{j-i} \right)
				-  m_j.
		\label{eq:lintran-components}
	\end{equation}
	Putting $\vec{X} \equiv (X_1,\ldots,X_\order)^T$,
	$\vec{Y} \equiv (Y_1,\ldots,Y_\order)^T$ and
	$\vec{m} \equiv (m_1,\ldots,m_\order)^T$, this can be written as a vector equation,
	\begin{equation}
		\vec{Y} = A \vec{X} - \vec{m},
		\label{eq:lintran}
	\end{equation}
	where the matrix $A$ is lower triangular
	with ones along the main diagonal:
	\begin{equation}
		A = \begin{pmatrix}
						1 & 0 & 0 &  \ldots & 0 \\
						-\arc_1 & 1 & 0 & \ldots & 0 \\
						-\arc_2 & -\arc_1 & 1 & \ldots & 0 \\
						\vdots & \vdots & \vdots & \ddots & \vdots \\
						-\arc_{\order-1} & -\arc_{\order-2} & -\arc_{\order-3} & \ldots & 1
		\end{pmatrix}
	\end{equation}
	Equation \eqrf{lintran} implies that $H(\vec{Y}) = H(\vec{X}) + \log \det{A}$, but 
	since $A$ has the above form, $\det{A}=1$ and so $H(\vec{X})=H(\vec{Y})$.
	Substituting \eqrf{cond-iir} and \eqrf{offsets} into \eqrf{lintran-components}:
	\begin{equation}
		 Y_j = U_j + \arc_j X_0.
	\end{equation}
	Expanding $X_0$ using \eqrf{cond-iir0} and writing in vector form,
	\begin{equation}
		\vec{Y} = \vec{U} - U_0 \vec{a}- m_0\vec{a},
	\end{equation}
	where $\vec{U}$ is a spherical Gaussian random vector and $\vec{a}$ is a constant
	vector with components $a_i = -\arc_i$.
	The determinant of the covariance of $\vec{Y}$
	and hence the entropy $H(\vec{Y})$ can be found by exploiting the spherical symmetry
	of $\vec{U}$ and rotating into a frame of reference in which $\vec{a}$ is aligned
	with first coordinate axis and therefore has the components 
	$(\norm{\vec{a}}, 0, \ldots)$. In this frame of reference, 
	the covariance matrix of $\vec{Y}$ is
	\begin{equation*}
		\iftwocol{}{
		\innvar^2 I +
		\innvar^2 \begin{pmatrix}
				\norm{\vec{a}}^2 & 0 &  \ldots & 0 \\
				0 & 0 &  \ldots & 0 \\
				\vdots & \vdots & \ddots & \vdots \\
				0 &0 &  \ldots & 0
		\end{pmatrix}
		=}
		\innvar^2 \begin{pmatrix}
				1+\norm{\vec{a}}^2 & 0 &  \ldots & 0 \\
				0 & 1 &  \ldots & 0 \\
				\vdots & \vdots & \ddots & \vdots \\
				0 &0 &  \ldots & 1
		\end{pmatrix},
	\end{equation*}
	and therefore its determinant is just 
	$\innvar^{2\order}(1+\norm{\vec{a}}^2)$.
	This gives us the entropy of $\vec{Y}$ and hence of $\vec{X}$:
	\begin{equation}
		H(\vec{X}) = H(\vec{Y}) = \half \log (2\pi e\innvar^2)^\order (1+\norm{\vec{a}}^2).
		\label{eq:cond-block-entro}
	\end{equation}
	By construction, $H(\vec{X}) = H(X_{1:\order}|\ev(X_{-\order:-1}=x_{-\order:-1}))$; that is,
	the entropy of $X_{1:\order}$ conditioned on the particular observed values
	$x_{-\order:-1}$, but since \eqrf{cond-block-entro} is independent of those values, 
	we may conclude that $H(X_{1:\order}|X_{-\order:-1}) = H(\vec{X})$ and substitute \eqrf{cond-block-entro}
	and \eqrf{entro-ar} into \eqrf{pir-markov-entro} to obtain
	\begin{equation}
		\begin{split}
			b_\mu &= H(X_{1:\order}|X_{-\order:-1}) - Nh_\mu \\
			&= \half \log (2\pi e\innvar^2)^\order (1+\norm{\vec{a}}^2) - \half \order \log 2\pi e \innvar^2.
		\end{split}
		\label{eq:pir-ar-pre}
	\end{equation}
	Simplified and expressed in terms of the original filter coefficients $\arc_k$, 
	\begin{equation}
					b_\mu = \half \log \left(1+\sum_{k=1}^\order \arc_k^2\right).
		\label{eq:pir-ar}
	\end{equation}

	Let us now consider the relationship between the PIR and power spectral density.
	For an autoregressive process, the PSD can be computed directly from the 
	prediction coefficients
	via the filter transfer function, which
	is the z-transform of the filter impulse response.
	If we set $a_0=1$ and $a_k = -\arc_k$ for $1\leq k \leq \order$, and
	temporarily reuse the symbol $H(\cdot)$ to denote the transfer function, then,
	\begin{equation}
		H(z) = \frac{1}{1 - \arc_1 z^{-1} \ldots - \arc_\order z^{-\order}}
		= \frac{1}{\sum_{k=0}^\order a_k z^{-k}},
		\label{eq:transfer-fn}
	\end{equation}
	and $S(\omega) = \innvar^2 \abs{H(e^{i\omega})}^2$. Since all the coefficients are assumed
	real, this gives
	\begin{align*}
		S(\omega) &= \frac{\innvar^2}{\sum_{k=0}^\order a_k e^{i\omega k} \sum_{j=0}^\order a_j e^{-i\omega j}},
		 \\ &= \frac{\innvar^2}{\sum_{k=0}^\order\sum_{j=0}^\order a_k a_j e^{i\omega(k- j)}}.
	\end{align*}
	Now, consider the integral of $\innvar^2/S(\omega)$  over one cycle of $\omega$
	from $-\pi$ to $\pi$:
	\begin{align*}
		\int_{-\pi}^\pi \frac{\innvar^2}{S(\omega)} \dd\omega 
		 &= \int_{-\pi}^\pi \sum_{k=0}^\order\sum_{j=0}^\order a_k a_j e^{i\omega(k- j)} \dd \omega \\
		 &= \sum_{k=0}^\order\sum_{j=0}^\order a_k a_j \int_{-\pi}^\pi e^{i\omega(k- j)} \dd \omega \\
		 &= \sum_{k=0}^\order\sum_{j=0}^\order a_k a_j 2\pi \delta_{jk} 
		= 2\pi \sum_{k=0}^\order a_k^2.
	\end{align*}
	Referring back to \eqrf{pir-ar}, this shows that, for $\AR(\order)$ processes at least, 
	the predictive information rate can be expressed in terms of the power spectrum $S(\omega)$ as
	\begin{equation}
		b_\mu = \half \log \specint{\frac{\innvar^2}}.
	\end{equation}
	Substituting in \eqrf{inn-var-psd} for $\innvar^2$, we get
	\begin{equation*}
		b_\mu = \frac{1}{2} \left( \specint{\log}  + \log \specint{\frac{1}}\right).
	\end{equation*}
	As a final step, this can be written entirely in terms of the \emph{inverse} power spectrum,
	in an expression which is an exact parallel of \eqrf{mir-sfm}:
	\begin{equation}
		b_\mu = \frac{1}{2} \left( \log \specint{\frac{1}} - \specint{\log\frac{1}} \right),
		\label{eq:pir-psd}
	\end{equation}
	exposing an intriguing duality between the multi-information and predictive information
	rates on the one hand, 
	and Gaussian processes whose power spectra are mutually inverse on the other.
	A similar duality was noted by \citet{AbdallahPlumbley2012} in relation
	to the multi-information and the binding information in finite sets of discrete-valued
	random variables. 
	Although derived for finite-order autoregressive process, we conjecture that
	\eqrf{pir-psd} may be valid for any Gaussian process for which the required intergrals
	exist, and return to this topic later, in our analysis of moving-average processes.

	\subsection{Residual or erasure entropy rate}
	Since the erasure entropy rate is $r_\mu = h_\mu -b_\mu$,
	we can write $r_\mu$ in terms of the power spectrum as follows:
	\begin{equation}
		r_\mu = \frac{1}{2} \left( \log 2\pi e - \log \specint{\frac{1}} \right).
	\end{equation}
	This concurs with the results of \citet{VerduWeissman2006}, which are presented
	there without proof. We outline
	a skeleton proof later in \secrf{ma-pir}.

	\section{Autoregressive Examples}
	\label{s:ar-examples}
	
	Here we compute $b_\mu$ and $\rho_\mu$ for some simple cases to get
	feel for their range of variation.
	In all cases, the processess
	are normalised to unit variance, so that $\ac_0 = \expect{X_t^2} = 1$ for all $t$ and
	the marginal entropy $H(X_t) = \half\log 2\pi e$ and therefore
	$\rho_\mu + b_\mu = H(X_0)$ is constant.

	\subsection{$\AR(1)$ processes}
	\label{s:ar1}

	\begin{fig}{ar1-info}
		\colfig[\iftwocol{0.9}{0.65}]{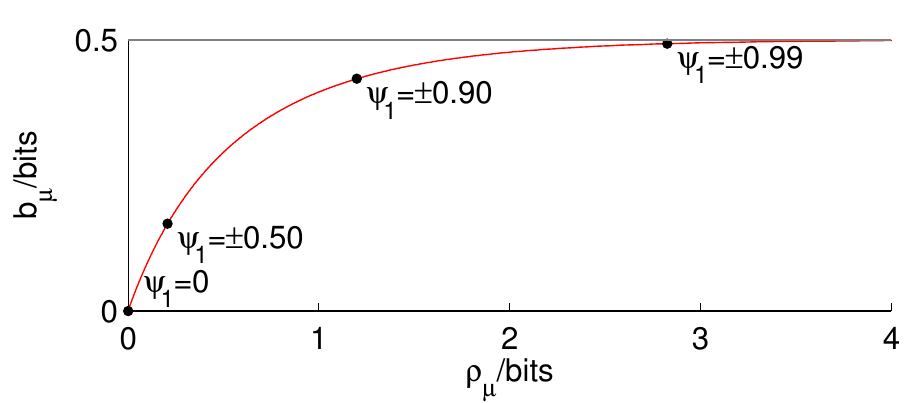}
		\caption{Multi-information rate $\rho_\mu$ and predictive information rate $b_\mu$
		for $\AR(1)$ processes with prediction coefficient
		$\arc_1$. The grey line is the asypmtote at $b_\mu = \half$ a bit.}
	\end{fig}

	The simplest case we can consider is that of the $\AR(1)$ processes, which, given
	the variance constraint, form a one-dimensional family parameterised by the
	prediction coefficient $\arc_1$. The generative equation is
	\begin{equation}
		X_t = U_t  + \arc_1 X_{t-1}.
		\label{eq:ar1}
	\end{equation}
	The process will be stationary only if the corresponding IIR filter is stable, 
	which requires that $\abs{\arc_1} < 1$. Multiplying $\eqrf{ar1}$ by 
	$X_{t-k}$ and taking expectations yields
	\begin{equation*}
		\expect{X_t X_{t-k}} = \expect{U_t X_{t-k}}  + \arc_1 \expect{X_{t-1} X_{t-k}},
	\end{equation*}
	from which we obtain the Yule-Walker equations relating the autocorrelation
	sequence $\ac_k= \expect{X_t X_{t-k}}$ and the prediction coefficient $\arc_1$:
	\begin{equation}
		\begin{split}
		\ac_0 &= \innvar^2 + \ac_1 \arc_1, \\
		\ac_1 &= \ac_0 \arc_1.
		\end{split}
		\label{eq:yule-walker}
	\end{equation}
	The variance constraint means that $\ac_0=1$ and so we find that $\innvar^2 = 1 - \arc_1^2$.
	Since $\rho_\mu = H(X_0) - h_\mu$ and $H(X_0)$ is fixed at $\half\log 2\pi e$, we obtain the 
	following results: 
	\begin{align}
		\rho_\mu &= - \half \log (1-\arc_1^2), \\
		b_\mu &= \half \log (1+\arc_1^2). \label{eq:ar1-pir}
	\end{align}
	Given the stability constraints on $\arc_1$, both quantities are minimised
	when $\arc_1=0$, which corresponds to $X$ being a unit-variance white noise
	sequence. Both $\rho_\mu$ and $b_\mu$ increase as $\arc_1 \tends \pm 1$:
	the multi-information rate diverges while the PIR tends to $\half\log 2$ nats or
	half a bit. As $\arc_1 \tends 1$, the process becomes Brownian noise,
	whose sequence of first differences are white noise. 
	The marginal variance constraint means that the innovation variance $\innvar^2$
	simultaneously tends to zero. However, since $b_\mu$ is invariant to rescaling
	of the processes, the PIR of any (discrete-time) Brownian noise
	can be taken to be $0.5$ bits per sample.
	If $\arc_1 \tends -1$, the
	process no longer looks like Brownian noise, but can be obtained from one 
	by reversing the sign of every other sample, that is, by applying the map
	$X_t\mapsto (-1)^t X_t$. 

	\subsection{$\AR(2)$ processes}
	\label{s:ar2}

	\begin{fig}{ar2-info}
		\iftwocol{%
			\colfig[0.83]{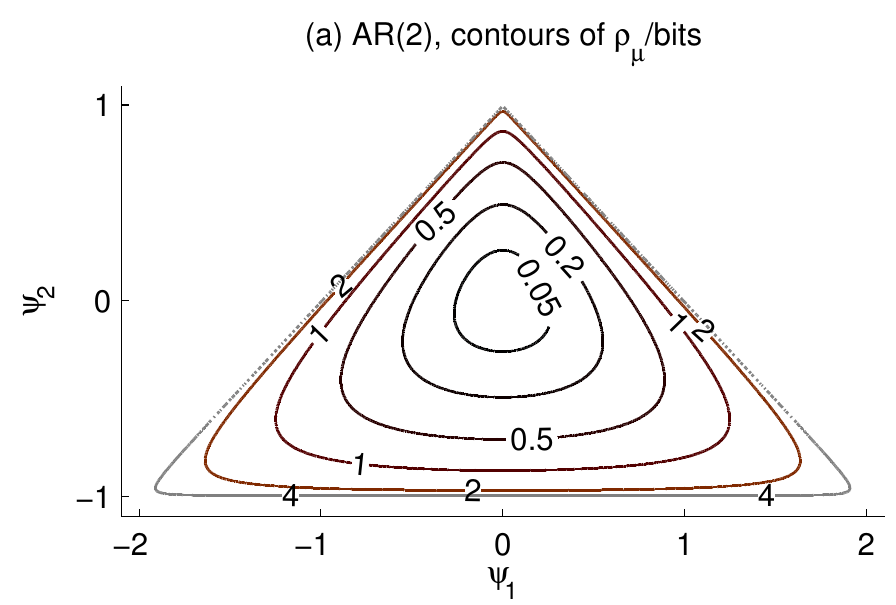}\\[1em]
			\colfig[0.83]{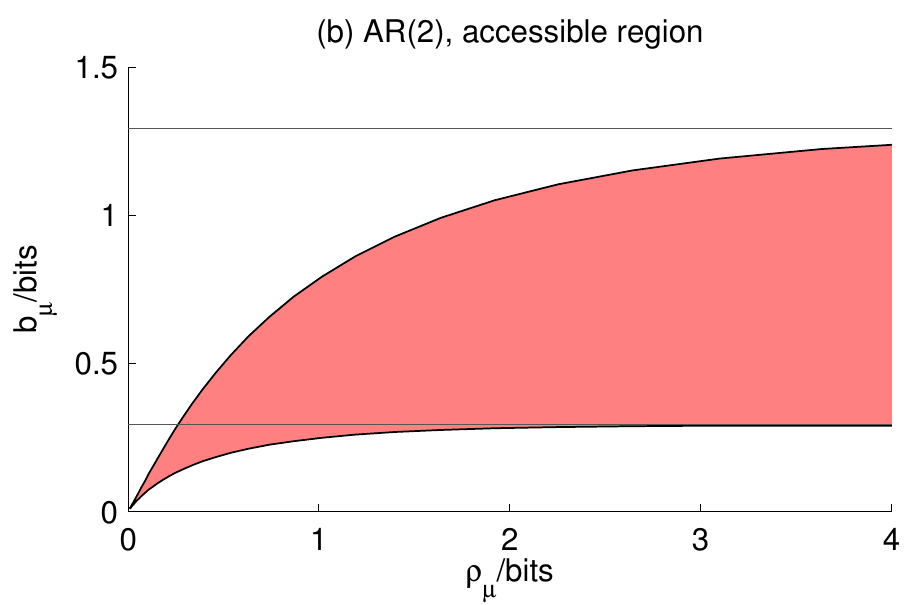}
		}{
			\colfig[0.49]{figs/ar2-rho-contours} \colfig[0.49]{figs/ar2-rho-b}
		}
		\caption{(a) Multi-information rate $\rho_\mu$ for $\AR(2)$
		processes parameterised by prediction coefficients 
		$\arc_1$ and $\arc_2$. Numerically computed contours of $\rho_\mu$ over a range from 
		$0.05$ to $4$ are shown. The interior of the triangle $(1,0), (2,-1), (-2,-1)$ 
		is the region of stability of the process, and $\rho_\mu$ diverges to infinity at
		its edges. Contours of $b_\mu$ are not shown as they are simply circles centred
		at the origin. (b) Accessible region (shaded) of $(\rho_\mu,b_\mu)$ pairs corresponding
		to interior of triangle in (a), computed numerically from contours at many values
		of $\rho_\mu$. Upper and lower grey lines are upper and lower
		asymptotes at $\half \log_2 6 \approx 1.29$ bits and 
		$\half\log \tfrac{3}{2} \approx 0.29$ bits respectively.}
	\end{fig}

	Second-order processes can be tackled in much the same way. In this case,
	the generative equation is
	\begin{equation}
		X_t = U_t  + \arc_1 X_{t-1} + \arc_2 X_{t-2},
		\label{eq:ar2}
	\end{equation}
	and the Yule-Walker equations are
	\begin{equation}
		\begin{split}
		\ac_0 &= \innvar^2 + \ac_1 \arc_1 + \ac_2 \arc_2, \\
		\ac_1 &= \ac_0 \arc_1 + \ac_1 \arc_2, \\
		\ac_2 &= \ac_1 \arc_1 + \ac_0 \arc_2.
		\end{split}
		\label{eq:yule-walker2}
	\end{equation}
	A little algebra eventually yields
	\begin{equation}
		\ac_0 = \frac{\innvar^2 (1 - \arc_2)}{
						1 - \arc_1^2 - \arc_1^2 \arc_2 - \arc_2 - \arc_2^2 + \arc_2^3}
		\end{equation}
		and therefore
	\begin{align}
		\rho_\mu &= \half \log \frac{1 - \arc_2}{
						1 - (\arc_1^2 + \arc_2)(1 + \arc_2) + \arc_2^3}, \\
			b_\mu &= \half\log ( 1+ \arc_1^2 + \arc_2^2).
		\end{align}
	\Figrf{ar2-info} illustrates how $\rho_\mu$ and $b_\mu$ vary with
	$\arc_1$ and $\arc_2$. The PIR is maximised when $\arc_1=\pm 2$ and
	$\arc_2=-1$, which corresponds to a transfer function $H(z)$ (as in equation
	\ref{eq:transfer-fn}) with two poles in the same place at $z=\pm 1$.
	Similar to the $\AR(1)$ case, as $(\arc_1,\arc_2)$ approaches $(2,-1)$, 
	$\rho_\mu$ diverges and the process becomes non-stationary, but insteading of becoming
	Brownian noise, it becomes the cumulative sum of a Brownian noise: a quick check
	will verify that the sequence of \emph{second} differences is white noise.

	\subsection{$\AR(N)$ processes with random poles}
	\label{s:arn}

\comment{
	\begin{fig}{arn-scat}
		\colfig[\iftwocol{0.83}{0.6}]{figs/arn-scat}
		\caption{Multi-information rate and PIR for 5000 AR(8) systems generated by randomly
		sampling poles (non-uniformly and in conjugate pairs) inside the unit circle. The grey line is a conjectured upper asymptote
		at $\half \log_2 12870 \approx 6.83$ bits, which corresponds to all 8 poles at $z=1$.}
	\end{fig}
}
	In our final example, we take a look at higher-order autoregressive processes.
	In order to ensure that we consider only stable processes, we generate them by randomly
	sampling their poles inside the unit circle in the complex plane. From the poles
	we compute the prediction coefficients by expanding the factorised form
	of the transfer function, which, for $\order$ poles at $\zeta_1,\ldots,\zeta_\order$, is
	\begin{equation}
		H(z) = \frac{z^\order}{\prod_{k=1}^\order (z-\zeta_k)}.
	\end{equation}
	The autocorrelation sequence $\ac_k$ can be computed from $\innvar^2$ and the prediction 
	coefficients using what is essentially a generalisation of the methods used in \secrf{ar1} 
	and \secrf{ar2} (
	as implemented in the \textsc{MATLAB}
	function \texttt{rlevinson}). From $\ac_0/\innvar^2$ we compute $\rho_\mu$.
	The points cover a region qualitatively similar to that shown in \figrf{ar2-info}(b), but
	with different upper and lower asymptotes. Our initial
	investigations suggest that the upper limit of $b_\mu$ is approached if all the
	poles approach $1$ or $-1$, at which point the prediction coefficients are the binomial
	coefficients and are easily computed. For example, at $\order=8$, we obtain 
	$b_\mu < \half \log 12870$. As with $\AR(1)$ and $\AR(2)$, the resulting processes
	are such that the \nth{\order} differences of the sequence are white noise, but
	since the variance of the innovations tends to zero, the processes themselves
	appear increasingly smooth and are dominated by low frequencies.

	\section{Moving-average Processes}
	\label{s:moving-average}

	A moving-average Gaussian process of order $\order$ is a real-valued random process
	$(X_t)_{t\in\integers}$ such that 
	\begin{equation}
		X_t = \sum_{k=0}^\order b_k U_{t-k},
		\label{eq:fir}
	\end{equation}
	where the $U_t$ form a sequence of independent 
	Gaussian random variables with zero mean and variance $\innvar^2$. 
	Thus, a realisation of the process $X$ is the result of applying an order-$\order$
	finite impulse response (FIR) filter with coefficients $b_k$ to a realisation of the sequence formed
	by the $U_t$. 
	The class of such processes is known as $\MA(\order)$.
	Without loss of generality, we may assume $b_1=1$, since any overall scaling
	of the process can be absorbed into $\innvar^2$. We may also assume 
	that none of the roots of the filter transfer function polynomial
	$B(z)=\sum_{k=0}^N b_k z^{-k}$ are outside the unit disk in the complex plane, by the following argument:
	assuming $b_0=1$, $B(z)$ can be expressed in terms of its $N$ roots $\beta_1,\ldots,\beta_N$ as 
	\begin{equation}
		B(z) = \frac{1}{z^N}\prod_{k=1}^N (z - \beta_k).
	\end{equation}
	The spectral density at angular frequency $\omega$ is therefore
	\begin{equation}
		S(\omega) = \innvar^2 \abs{B(e^{i\omega})}^2 = \innvar^2 \prod_{k=1}^N \abs{e^{i\omega} - \beta_k}^2.
		\label{eq:mapsd}
	\end{equation}
	The Gaussian process is uniquely determined by giving either its autocorrelation sequence or its
	spectral density function. If we move any of the roots $\beta_k$ without changing the value of
	$S(\omega)$ for any $\omega\in\reals$, the FIR filter may be different but the process itself will
	be remain unchanged. Suppose one of the roots is at $\zeta$ and $\abs{\zeta}>1$. 
	Its contribution to the PSD is a factor of
	\begin{align*}
		\abs{e^{i\omega} - \zeta}^2
			&= \abs{ \zeta e^{i\omega} ( 1/\zeta - e^{-i\omega})}^2 
			= \abs{\zeta}^2 \abs{ e^{i\omega} - \bar{\zeta}}^2,
	\end{align*}
	where $\bar{\zeta} = 1/\zeta^*$ is the reciprocal of the complex conjugate of $\zeta$ and
	hence inside the unit disk. Thus, the root $\zeta$ can be replaced with $\bar{\zeta}$ and
	the only effect on the PSD is the introduction of the constant factor $\abs{\zeta}^2$, which
	can be absorbed into $\innvar^2$.
	In this way, all the roots of $B(z)$ that are outside the unit disk can be moved inside
  without changing the statistical structure of the process.
	Noting that \eqrf{fir} can be written as 
	\begin{equation}
		U_t = X_t - \sum_{k=1}^N b_k U_{t-k},
		\label{eq:ma-as-iir}
	\end{equation}
	we see that
	the sequence $(\ldots,U_{t-1},U_t)$ can be computed from the 
	sequence $(\ldots,X_{t-1},X_t)$ via
	a stable IIR filter with the transfer function $1/B(z)$. These properties will be useful
	when we try to determine the process information measures of the $\MA(N)$ process.

	\begin{fig}{ma-gm}
		\begin{tikzpicture}[->]
			\def\cn(#1,#2) {\node[circle,draw,inner sep=0.2em] (#1#2) {$#1_#2$};}
			\def\dn(#1) {\node[circle,inner sep=0.2em] (#1) {$\cdots$};}
			\def\rl(#1,#2) {\draw (#1) -- (#2);}
			\def\cl(#1,#2) {\draw (#1) to [bend right,looseness=1.2] (#2);}
			\matrix[row sep=1.4em, column sep=1.9em]{
				\dn(UH) & \cn(U,1) &  \cn(U,2) & \cn(U,3) & \cn(U,4) &  \cn(U,5) & \dn(UT) \\
				\dn(XH) & \cn(X,1) & \cn(X,2) & \cn(X,3) & \cn(X,4) & \cn(X,5) & \dn(XT) \\
			};
			\rl(UH,X1) 
			\rl(U1,X1) \rl(U1,X2)
			\rl(U2,X2) \rl(U2,X3)
			\rl(U3,X3) \rl(U3,X4)
			\rl(U4,X4) \rl(U4,X5)
			\rl(U5,X5) \rl(U5,XT)
		\end{tikzpicture}
		\caption{Graphical model for an $\MA(1)$ first order moving-average Gaussian process. The $X_t$ are
		the observed, real-valued random variables, while the $U_t$ are unobserved. Each
		$X_t$ is a deterministic (linear) function of its parents.}
	\end{fig}

	The first thing to note about this model is that it is does not have the
	Markov conditional independence structure of the $\AR(N)$ model. Consider
	the graphical model of an $\MA(1)$ process depicted in \figrf{ma-gm}: even though
	$X_2$ and $X_4$ are \emph{marginally} independent (since their parent node sets are
	disjoint and independent), they become \emph{conditionally} dependent if
	$X_3$ is observed, because the known value of $X_3$ means that $U_2$ and $U_1$
	become functionally related. The same argument applies if an arbitrarily long
	sequence $X_{1:\len}$ is observed: in this case, $X_0$ and $X_{\len+1}$ become conditionally
	dependent given $X_{1:\len}$. 
	This lack of any finite-order Markov structure means that the 
	measures $h_\mu$, $\rho_\mu$ and $b_\mu$ cannot be computed from the joint
	distribution of any finite segment of the sequence, say $X_{-\len:\len}$, as we did
	in \secrf{ar}, but can be obtained
	by using spectral methods to analyse the covariance structure in the limit $\len\tends\infty$.
	From \eqrf{fir}, we obtain the autocorrelation sequence
	\begin{equation}
		\begin{split}
			\ac_m = \expect{X_t X_{t-m}} &= \sum_{k=0}^N \mac_k U_{t-k} \sum_{j=0}^N \mac_j U_{t-m-j} 
			 \\ &=
			  \sum_{k=0}^N \sum_{j=0}^N \mac_k  \mac_j \innvar^2\delta_{k,m+j} 
				\iftwocol{\\&}{}
				= \innvar^2 \sum_{k=m}^{N} \mac_k  \mac_{k-m},
		\end{split}
		\label{eq:maac}
	\end{equation}
	which is non-zero for at most $2N+1$ values of $m$, from $-N$ to $N$. Hence, the
	covariance matrix $R=\expect{\vX\vX\tran}$ of the multivariate Gaussian 
	$\vX \equiv (X_{-\len},\ldots,X_\len)$, when $\len>N$, will be a \emph{banded} toeplitz matrix.
	For example, for an $\MA(1)$ process it will be
	\begin{equation}
		\begin{pmatrix}
			\ac_0  & \ac_1 & \cdots & 0 \\
			\ac_1  & \ac_0 &         &   \\
			\vdots &       &  \ddots & \ac_1 \\
			0      &       &  \ac_1 & \ac_0 
		\end{pmatrix}.
		\label{eq:banded}
	\end{equation}

	\subsection{Entropy rate}
	In the case of $\MA$ processes and with our assumption that roots of the transfer function
	are not outside the unit disk, the Kolmogorov-Sinai entropy \eqrf{kge} 
	can be evaluated exactly by substituting
	in \eqrf{mapsd} and using Jensen's formula, 
	which gives $\int_{-\pi}^\pi \log \abs{e^{i\omega} - \zeta} \dd \omega = 0$ if $\abs{\zeta}\leq 1$:
	\begin{align*}
		\int_{-\pi}^\pi \log S(\omega) \dd \omega &= 
		\int_{-\pi}^\pi \log \innvar^2 \prod_{k=1}^N \abs{e^{i\omega}-\beta_k}^2 \dd \omega 
		\\ &= 
		\log \innvar^2 + 2 \sum_{k=1}^N \int_{-\pi}^\pi \log \abs{e^{i\omega}-\beta_k} \dd \omega.
		\\ &= \log \innvar^2,
	\end{align*}
	and hence
	\begin{equation}
		h_\mu = \half \log 2\pi e \innvar^2.
	\end{equation}
	This is consistent with our earlier observation that the innovations up to and including time $t$ can be
	computed from the observations up to time $t$ by IIR filtering the observations: in this case,
	the conditional variance of the next observation is just the variance of $b_0 U_{t+1}$, which is $\innvar^2$.

	\subsection{Multi-information rate}
	From \eqrf{fir} and \eqrf{maac}, the marginal variance is 
	$\expect{X_t^2} = \ac_0 = \innvar^2\sum_{k=0}^N b_k^2$,
	so, with $b_0=1$, and $\rho_\mu = H(X_t) - h_\mu$,  the multi-information rate is
	\begin{equation}
		\rho_\mu = \half \log \left( 1 + \sum_{k=1}^N b_k^2\right),
		\label{eq:mir-ma}
	\end{equation}
	which is in agreement with Ihara's result \citep[\S 2.2]{Ihara1993}. Note that this is dual to the
	result obtained for the predictive information rate in $\AR(N)$ processes \eqrf{pir-ar}, in that
	the FIR filter coefficients $b_k$ have taken the place of the IIR filter coefficients $\arc_k$
	or $a_k$.
		 
	\subsection{Predictive information rate}
	\label{s:ma-pir}
	The PIR can be obtained from the erasure entropy rate $r_\mu$ using the relation
	$b_\mu = h_\mu - r_\mu$. \citet{VerduWeissman2006} state without proof that the erasure
	entropy rate of a Gaussian process with power spectral density $S(\omega)$ is
	\begin{equation}
		r_\mu = \half \log 2 \pi e - \half\log \left(\specint{\frac{1}} \right),
		\label{eq:gp-erasure}
	\end{equation}
	which, in combination with \eqrf{kge}, yields
	\begin{equation}
		b_\mu = \frac{1}{2}\left( \specint{\log} + \log \specint{\frac{1}} \right),
	\end{equation}
	which agrees with the expression we obtained earlier for $\AR$ processes.
	A skeleton of a proof of \eqrf{gp-erasure} can be obtained by considering 
%
	the limit 
	of $H(X_0|X_{-\len:-1},X_{1:\len})$ as $\len\tends\infty$.
	Let $\vX\oflen$ be the random vector $(X_{-\len},\dots,X_\len)$ with covariance
	matrix $R\oflen$ constructed from the autocorrelation sequence as shown previously
	in \eqrf{maac} and \eqrf{banded}.
	If $K\uplen = R\oflen^{-1}$ is the corresponding precision matrix, the
	probability density function $p\oflen : \reals^{2\len+1}\to\reals$ for $\vX\oflen$ 
	is the multivariate Gaussian:
	\begin{equation}
		p\oflen(\x) \propto \exp{ -\half {\x}^\top K\uplen \x}. 
	\end{equation} If we index the elements of $\x$ and $K^\len$ starting with $-\len$ and running through $0$
	to $\len$, then it can easily be shown by examing the functional dependence of $p\oflen(\x)$
	on $x_0$ that the conditional density of the central variable $X_0$ given
	given the values of all the others is Gaussian with variance $1/K^\len_{00}$, and hence the conditional
	entropy is $H(X_0|X_{-\len:-1},X_{1:\len}) = - \half \log 2\pi e K\uplen_{00}$.
	Now, since $R\oflen$ is real and symmetric, it will have $2\len+1$ orthogonal eigenvectors with
	real eigenvalues, and $K\uplen$ can be represented in terms these as 
	\begin{equation}
		K\uplen_{jk} = \sum_{n=-\len}^\len r_n^{-1} V_{jn} V_{kn}^*,
		\label{eq:precision-mat}
	\end{equation}
	where $V_{jn}$ is the \nth{j} component of the \nth{n} eigenvector with eigenvalue $r_n$.
	If $R\oflen$ had been \emph{circulant} as well as toeplitz, its eigenvectors would have been 
	complex exponentials of the form $V_{jn} = e^{-2\pi i j n/\len}/\sqrt{2\len+1}$, in which
	case, substitution into \eqrf{precision-mat} would yield $\sum_n r_n^{-1}/(2\len+1)$ 
	for all the diagonal elements. Instead, the standard approach \citep{Gray2006} is to
	construct two infinite sequences of matrices with are \emph{asymptotically equivalent}.
	The first sequence consists of covariance matrices $R\oflen$ as $\len$ increases, \ie,
	$R_1, R_2$, \etc The second is a sequence of circulant approximations of the $R\oflen$.
	As $\len\tends\infty$, the sequences converge to each other (in the weak norm sense) and
	many properties of the $R\oflen$ converge to those of their circulant approximations.
	This does \emph{not} prove that all diagonal elements of the inverse $R\oflen^{-1}$ converge
	in this way, and indeed, we would not expect them to for the extremal elements such
	as $K^\len_{\len\len}$ as this would be inconsistent with the result for the entropy rate.
	However, numerical results suggest that for `central' elements $K^\len_{jj}$ such that
	both $j+\len$ and $\len-j$ tend to infinity as $\len$ tends to infinity, we can assume
	that the values do converge as expected. In particular, for the middle element, we suppose that
	\begin{equation}
		\lim_{\len\tends\infty}K\uplen_{00} = \frac{1}{2\len +1 }\sum_{n=-\len}^\len r_n^{-1}.
	\end{equation}
	This remains to be proved, but if we accept it, then by Szeg\"{o}'s theorem \cite{Gray2006}, 
	which applies to such functions of the eigenvalues of a toeplitz 
	matrix, this converges to an integral expressed in terms of the spectral density function:
	\begin{equation}
		\lim_{\len\tends\infty}   \frac{1}{2\len+1} \sum_{n=-\len}^\len \frac{1}{r_n}
		= \specint{\frac{1}},
	\end{equation}
	so we obtain the expected expression for the erasure entropy rate
	\begin{equation}
		r_\mu = \half \log 2 \pi e - \half\log \left(\specint{\frac{1}} \right).
	\end{equation}

	\section{Moving-average Examples}
	\label{s:ma-examples}

	\def\second{$2^\text{nd}$\xspace}
	\def\first{$1^\text{st}$\xspace}
	\def\third{$3^\text{rd}$\xspace}

	The simplest non-trivial moving-average process that we can consider is the $\MA(1)$
	process
	\begin{equation}
		X_t = U_t + b_1 U_{t-1},
		\label{eq:ma2}
	\end{equation}
	where the sole parameter $b_1$ satisfies $\abs{b_1}<1$, according to the assumptions
	described at the beginning of \secrf{moving-average}. Using \eqrf{mir-ma}, we find that
	$\rho_\mu = \half \log (1+b_1^2)$, which is dual to the result \eqrf{ar1-pir} obtained for
	the PIR of the $\AR(1)$ process obtained by inverting the spectrum
	of this $\MA(1)$ process. The transfer function of the two-tap FIR filter from 
	$U$ to $X$ is $H(z) = 1 + b_1 z^{-1}$. If we define $\bar{H}(z) = 1/H(z) = 1/(1 + b_1 z^{-1})$,
	we can see that $\bar{H}(z)$ is the transfer function of the \first order IIR filter
	associated with an $\AR(1)$ process, where $b_1$ plays the role
	of the prediction coefficient. Clearly, the spectrum of this process, call it $\bar{X}$, 
	will be the inverse of the original process, and we can use the results of 
	\secrf{ar}, along with the duality relationship we observed relating the multi-information
	and predictive information rates, to compute the multi-information and predictive information
	rates of the moving-average process $X$. Referring back to \secrf{ar-examples}, we obtain
	\begin{align}
		\rho_\mu &= \half \log (1+b_1^2), \label{eq:ar1-pir} \\
		b_\mu &= - \half \log (1-b_1^2).
	\end{align}
	Rather than repeat the process of illustrating these equations, we refer the reader
	back to \figrf{ar1-info}: the relationship is the same except for swapping the
	$\rho_\mu$ and $b_\mu$ axis labels and replacing $\arc_1$ with $b_1$.
	Indeed, the same reasoning can be applied to higher-order moving-average processes,
	so we can reuse figure \ref{f:ar2-info} for moving-average process
	by swapping $\rho_\mu$ and $b_\mu$, and replacing the prediction coefficients
	$\arc_k$ with the moving-average coefficients $b_k$.

	One implication of these results is that, even in the $\MA(1)$ process, the PIR approaches
	infinity as $b_1$ approaches $\pm 1$. In higher-order processes, the PIR diverges as the
	zeros of the transfer function approach the unit circle in the complex plane. In particular,
	the dual of the $\AR(N)$ process identified in \secrf{arn}, with all poles together at $1$ or $-1$,
	is an $\MA(N)$ process with all zeros at $1$ or $-1$, and maximises $\rho_\mu$ as $b_\mu$ diverges.
	With all zeros at $-1$, the coefficents of the corresponding FIR filter are the binomial 
	coefficients, and so as the order $N$ tends to infinity, the filter approximates a smoothing filter
	with a Gaussian impulse response.

	\section{Discussion and conclusions}

	We have found a closed-from expression for the predictive information
	rate in autoregressive Gaussian processes of arbitrary
	finite order, which is a simple function of the predictive coefficients.
	It can also be expressed as function of the power spectral density of the
	process in a form which we conjecture may apply to arbitrary Gaussian processes
	and not just autogressive ones. The functional form also suggests a
	duality between the PIR and multi-information rate, since the PIR of 
	a process with power spectrum $S(\omega)$ equals the multi-information
	rate of a process with the inverse power spectrum $1/S(\omega)$. 

	The fact that the stationary $\AR(1)$ and $\AR(2)$ processes maximising the
	PIR turn out to be, in the limit, Brownian motion and its (discrete time)
	integral is intruiging and perhaps counter-intuitive: in order to preserve
	finite variance, both process have vanishingly small innovations, with $\innvar^2$
	tending to zero as the limit is approached, and therefore `look smooth'.
	Indeed, as the order $\order$ is increased, the results of \secrf{arn} suggest
	that this pattern continues, with the PIR-maximising processes being increasingly
	`smooth' and having power spectra more and more strongly peaked at $\omega=0$.
	The PIR, originally proposed \cite{AbdallahPlumbley2009} as a potential 
	measure of complexity or `interestingness' (for which purpose it seems
	a plausible candidate, at least for discrete valued processes), is telling us that
	these very `smooth' Gaussian processes are somehow the most `interesting'.

	The difficulty is presented even more starkly in the case of moving-average processes,
	where the PIR is unbounded, and we are forced to conclude that a single observation
	can yield infinite information about the unobserved future. Once again, we find 
	that very `smooth' looking processes can have arbitrarily high predictive information
	rates.

	The reason for this, we suggest, lies in the assumption that variables in a
	real-valued random sequence can be observed  with \emph{infinite} precision. 
	Under
	these conditions, the tiny innovations observed in the unit-variance almost-Brownian noise
	of $\AR(1)$ when $\arc_1$ approaches 1 are just as measurable as the macroscopic
	innovations in the non-Brownian case and are significant and informative in a predictive
	sense, because every innovation is preserved into the infinite future in the
	form of an additive shift to all subsequent values in the sequence. 
	In addition, as soon
	as we have infinite precision measurements, we open the door to the possibility
	of infinite information; hence the divergence of $\rho_\mu$ and $b_\mu$ in these limiting cases.
	This rather un-physical
	situation can be remedied if we recognise that, in physically realisable systems, 
	the variables can only be observed
	with finite precision, either by explicitly modelling a quantisation error or by
	introducing some `observation noise', for example, by allowing infinite precision
	observations only of $Z_t = X_t + N_t$, where the $N_t$ are independent and Gaussian with
	some variance $\sigma_n^2$. In this case, each observation can only yield a finite
	amount of information about $X_t$, and it will no longer be possible to use infinitesimal
	variations to carry information about the future because they will be swamped by the observation
	noise. Recognising that what we are talking about here is essentially a hidden Markov model,
	we aim to establish these ideas on a more rigorous footing in future work.

	\begin{acknowledgments}
		\TheAcknowledgments
	\end{acknowledgments}
	\bibliography{all,c4dm,compsci}
\end{document}